\documentclass[conference]{IEEEtran}
\usepackage{cite}

\usepackage{booktabs}
\usepackage{array}

\ifCLASSINFOpdf
\usepackage[pdftex]{graphicx}
\else
\fi
\usepackage[cmex10]{amsmath}
\usepackage{tabulary}

\usepackage{algorithmic}

\usepackage{mdwmath}
\usepackage{mdwtab}

\usepackage[tight,footnotesize]{subfigure}

\usepackage{tikz}
\usetikzlibrary{positioning}

\usepackage{placeins}

\begin{document}
%
\title{Quasar Detection using Linear Support Vector Machine with Learning From Mistakes Methodology}  
\author{\IEEEauthorblockN{Aniruddh Herle,  \textit{Student Member, IEEE,} Janamejaya Channegowda,  \textit{Member, IEEE}, Dinakar Prabhu,  \textit{Member, IEEE}}}  
 
\maketitle
\begin{abstract} 
The field of Astronomy requires the collection and assimilation of vast volumes of data. The data handling and processing problem has become severe as the sheer volume of data produced by scientific instruments each night grows exponentially. This problem becomes extensive for conventional methods of processing the data, which was mostly manual, but is the perfect setting for the use of Machine Learning approaches. While building classifiers for Astronomy, the cost of losing a rare object like supernovae or quasars to detection losses is far more severe than having many false positives, given the rarity and scientific value of these objects. In this paper, a Linear Support Vector Machine (LSVM) is explored to detect Quasars, which are extremely bright objects in which a supermassive black hole is surrounded by a luminous accretion disk. In Astronomy, it is vital to correctly identify quasars, as they are very rare in nature. Their rarity creates a class-imbalance problem that needs to be taken into consideration. The class-imbalance problem and high cost of misclassification are taken into account while designing the classifier. To achieve this detection, a novel classifier is explored, and its performance is evaluated. It was observed that LSVM along with Ensemble Bagged Trees (EBT) achieved a 10x reduction in the False Negative Rate, using the Learning from Mistakes methodology.

\hspace*{-0.6cm} Keywords - Classification, machine learning,  support vector machines, quasars, astronomy

\end{abstract}

\IEEEpeerreviewmaketitle

\section{Introduction}
 One of the most important uses of Machine Learning (ML) in astronomy is for the  isolation of candidates within data streaming in from active telescopes for follow up analysis. Once a scientifically interesting object has been found, specialised telescopes, containing instruments designed specifically for those kinds of objects, are used to gather data related to objects of interest. This means that once a potential candidate is identified, another instrument is used to follow-up on it. To achieve this, the large amounts of data from certain telescopes must be screened and any object of scientific value must be identified quickly.\\
\indent A Quasar is an extremely rare class of astronomical objects that are abbreviated to QSO (Quasi-Stellar Object). They are extremely bright Active Galactic Nuclei in which a supermassive black hole is surrounded by extremely luminous in-falling matter. They are the most luminous persistent sources in space. They are also very compact, which means that they have a large power density. The energy is radiated across the entire electromagnetic spectrum, and they appear in the night sky with a star-like brightness, although the brightness variations and spectrum of the radiation emitted by these objects is very different from those of stars. \\
\indent These objects are very important scientifically. They can be used to study and interpret many scientific phenomena, like black-hole physics and ultra-high energy gamma rays \cite{Whiting2000}, which involve conditions of extreme gravity and energy densities that could never be re-created on Earth. Using quasars to calculate cosmological parameters has yielded very promising results, and researchers have also used quasars as distance indicators  \cite{Sulentic2014,Dultzin2020}. One of the most important aspects of Quasars is that they can be used as ‘standard candles’. This is due to the fact that their intrinsic luminosities can be calculated, and together with their apparent luminosity in the sky, can be used to determine their distance from us. This directly corresponds to how far back into the past they extend, meaning that they can be used to study the adolescent phases of the Universe's expansion, when it was only a billion years old. This in turn allows a better understanding of dark energy \cite{Risaliti2019}, which is one of the most perplexing and important questions faced by cosmologists today.

\section{Objective of the Paper}
 All the studies using quasars are data intensive. The accuracy and reliability of the results is dependent on the number of quasar candidates used for their calculations. The higher the number of samples used, the better the results will be. This can only be made possible using specialised classifiers designed to take into account the rarity of these objects. An important issue with existing datasets is the high class imbalance ratio, which is the ratio of the number of common objects to the number of rare objects, such as quasars. This is a major hurdle for supervised training techniques as the trained model becomes very biased to the majority class. This is because of the fact that the minority class is under-represented in the training data, due to its rarity. The effective handling of the class imbalance ratio is very important in astronomy. \\
\indent A large number of machine learning techniques have been developed to be used for working with astronomical data. Several key machine learning techniques in the fields of Astronomy, Astrophysics and Cosmology include Support Vector Machines \cite{Romano2006, Khramtsov2018}, Convolution Neural Networks \cite{Pearson2018}, Cascade Convolution Neural Networks \cite{Qahwaji2007}, Structure from Motion \cite{Elasmar2016}, and Deep Learning\cite{Mathuria2019, He2019}.\\ 
\indent Another aspect of the problem that needs to be taken into consideration is the cost of misclassifying a quasar as another class of more common objects. Given the rarity of these objects, it is important to minimise the number of candidates lost due to misclassification error, which occurs when the classifier places a quasar candidate into a class that it does not belong in. The False Negative Rate (FNR) represents the number of Quasars misclassified as a Non-Quasar. Hence, it is vital to reduce the FNR.\\
\indent This study takes into account the class-imbalance ratio and the high cost of losing a quasar candidate to misclassification, and attempts to implement a machine learning approach to identify quasars. A Linear SVM is trained on the Sloan Digital Sky Survey (SDSS) Data, and cascaded with an Ensemble Bagged Tree Algorithm to create a Quasar Classifier with a much reduced FNR. 

\section{Related Work}
Over the years, several attempts have been made to design classifiers for Astronomical data analysis. Mahabal et al. \cite{Mahabal2008} suggested a software methodology using Bayesian Networks for iterative probabilistic classification of astronomical sources found in large digital sky surveys. As much of the data in Astronomy is visual, computer vision techniques are also very popular. In Pearson et al. \cite{Pearson2018}, a Convolution Neural Network was developed to automatically detect gravitational lenses, called LensFinder, which occur when a massive astronomical object curves the surrounding space-time so much that they distort the path of light from more distant objects and act like a lens. The model segregates the input images into two classes, ones that contain lenses and ones that do not have lenses. After training LensFinder on the 420000 simulated images, the accuracy obtained was 98.19 ± 0.12\%. LensFinder averaged a false positive and false negative rate of approximately 1\%, with the FPR being slightly higher than the FNR. Work done by Makhija et al. \cite{Makhija2019} and Khramtsov et al. \cite{Khramtsov2019} focus on Tree methods for Quasar identification.\\
\indent Romano et al. \cite{Romano2006}  used Support Vector Machines to analyse large volumes of data to identify candidates for follow-up with specialized instruments. Data taken each night by the Nearby Supernova Factory (SNfactory), which is an international collaboration that studies Type Ia supernovae, must be filtered in such a way as to single out potential science targets. Supernovae are short-lived events that are very rare, occurring about a few times every million years, the study of which requires specialized instruments. To increase the recognition accuracy for unseen data, an incremental sampling approach was implemented, together with a selective training system where positive data was oversampled and negatives were under sampled. Implementing supervised learning, a new model is retrained using false negatives from the previous model’s outcome \cite{Romano2006}. The results of the test were highly optimistic, with final positive recognition rate of 99.5\% after the fourth iteration. Previously, the search for supernovae required 8 times more human working hours than after the implementation of the Support Vector Machines approach. \\
\indent SVMs have also been used in Khramtsov et al. \cite{Khramtsov2018} to classify objects into extragalactic or intergalactic objects. This was done using One-Class SVMs, as the objects are either extragalactic or not. The Sloan Digital Sky Survey Data Release was used for training the model. Representative learning, which is an algorithm that transforms input data into a feature space, was used to make it easier to isolate and train with features that represent the dataset well. In Khramstov et al. \cite{Khramtsov2018}, the One-Class SVM was applied to about 1.7 million objects in the sky, and the results showed an accuracy of 99.284\% for extragalactic objects.\\
\indent  In Heitz et al. \cite{Heitz2009}, several smaller computer vision problems were solved with different machine learning algorithms (like object detection, region labelling, geometric reasoning). Their framework organizes different classifiers into ‘tiers’. Lower tiers take inputs from the raw data as well as outputs of the higher tier classifiers. They report increments in accuracy of 7\% and 3\% for image categorization and segmentation tasks, respectively. Another way to cascade models would be to couple only the outputs of one model to the input of another, in such a way that the secondary model corrects any misclassification of the previous one. This has been shown to be effective in Bhatnagar et al. \cite{Bhatnagar2010}. Using Supernovae data, they showed that a Learning From Mistakes approach outperformed both Partitioning Ensemble (PARTEN) and Undersampling Majority Class (UMjC) methods on a large number of different datasets \cite{Bhatnagar2010}. The LFM framework gave better accuracies (95.78\%, 93\%, 94.87\%) and precision (57\%, 67\%, 84\%) values than PARTEN and UMjC frameworks when used for classifying different Supernovae into various sub-groups. 

\section{Dataset}
 The SDSS is a multi-spectrum and spectroscopic survey using a 2.5 m wide-angle telescope. This telescope operates out of the Apache Point Observatory, in New Mexico, USA. 
The 14th data release of the Sloan Digital Sky Survey \cite{Abolfathi2018} consists of 10,000 observations of the sky taken by the SDSS \cite{Blanton2017, Gunn2006}. This data release was part of the fourth phase of SDSS (SDSS-IV), containing observations through July 2016. Every observation is described by 17 feature columns and 1 class column which identifies it to be either a star, galaxy or quasar. Of these, only Right-Ascension, Declination (which indicate the position of the object), the flux magnitudes given by the Thuan-Gunn system \cite{Thuan1976}, and the redshift value are relevant for classification purposes. The data corresponds to three classes of objects: Stars, Galaxies and Quasars. The Stars and Galaxies classes have been merged together to form the Non-Quasi-Stellar Object class (NQSO). This results in the formulation of this problem as a two class classification problem (Quasar vs Non-Quasar).\\
\indent The Quasars observed in the SDSS are considered to be extremely far away which means that the redshift value associated with these quasars will be much higher than for stars and galaxies, with a value typically much greater than 0.1 \cite{Makhija2019}.This is important for building a classifier, as quasars are difficult to separate from Stars solely based on the flux values. This is because they are so far away that they appear as point sources, very similar to stars. \\
\indent Out of the 10,000 sets of datapoints, only 900 (9\%) are Quasars. The remaining 9100 belong to the combined class of both Stars and Galaxies. As can be seen, the number of Quasars in the data is relatively very low. It can be understood clearly that the class imbalance ratio is less than 1:10. This is attributed to the rarity of the Quasar. Due to this very rarity, losing a potential Quasar candidate for follow up study is very costly \cite{Khramtsov2019}. This poses a threat to straightforward classification methods, as the classifier will become biased towards the majority class, leading to very high False Negative Rates (FNR). 

\section{Analysis of Available Algorithms}
 In this paper, a Linear SVM based approach, coupled with an Ensemble Tree is implemented. Several of the best performing algorithms are discussed below.\\
\indent Matlab R2019B \cite{MATLAB:2019} was used to create and train these models. The Classification Learner Application was used to build and analyse the performance of several different types of algorithms. Several algorithms were trained on 60\% of the training data to classify the data points into Quasar and Non-Quasar, and the results are shown in Table \ref{table:1}.


 \begin{table}[htb!]
  \begin{center}
    \begin{tabular}{c c c c c} 
\textbf{Model} & \textbf{Accuracy} & \textbf{Total samples} & \textbf{Total} &\textbf{ Prediction} \\
 & (\%) & \textbf{classified} & \textbf{Misclassified} &  \textbf{Speed}\\
 &   & \textbf{correctly} &  & (Obs/sec)  \\       
  \hline
 Medium \\ Tree & 99.4 & 5991 & 9 & 130000  \\ 
 \hline
 Coarse \\ Tree & 99.5 & 5993 & 7 & 40000 \\ 
 \hline
 Quadratic \\ Discriminant & 99.4 & 5991 & 7 & 50000 \\ 
 \hline
 \textbf{Linear} \\ \textbf{SVM} & \textbf{99.7} & \textbf{5996} & \textbf{4} & 13000 \\ 
 \hline
 Quadratic \\ SVM & 99.6 & 5994 & 6 & 99000 \\ 
 \hline
 Cubic \\ SVM & 99.7 & 5995 & 5 & 57000 \\ 
 \hline
 Gaussian \\ SVM & 99.5 & 5992 & 8 & 84000 \\ 
 \hline
 Fine \\ KNN & 99.4 & 5991 & 9 & 19000 \\ 
 \hline
 Cosine \\ KNN & 99.3 & 5990 & 10 & 11000 \\ 
 \hline
 AdaBoosted \\ Trees \\ Ensemble & 99.5 & 5993 & 7 & 5200 \\ 
 \hline
 Bagged \\ Trees \\ Ensemble & 99.6 & 5994 & 6 & 10000 \\ 
 \hline
 RUS \\ Boosted \\ Trees \\ Ensemble & 99.6 & 5994 & 6 & 9200 \\ 
 \hline
   \end{tabular}
  \end{center}
\caption{Table of Model Performance}
\label{table:1}
\end{table}
\subsection{K-NN Classification}
This is a supervised learning approach that classifies a data point based on how its neighbours are classified. It is considered to be a ‘lazy learner’ as all computations are held off until the moment of the actual prediction, i.e a discriminative function is not learned. This computation involves calculating the Euclidean distance between two points in the feature space, and the $k$ smallest values correspond to the k-nearest neighbours. The majority label of these $k$ points is returned as the predicted class. An optimal value of $k$ must be chosen to achieve the highest accuracy.\\ 
\indent The Fine KNN algorithm works by choosing 1 as the value of $k$. Fine distinctions can be made on the input data in this case. In the Cosine KNN, the Cosine distance metric is used instead of using a Euclidean distance metric. This metric involves simply taking the dot product of two vectors. This method considers only the orientation of the vectors, and the dot product of one vector with another represents the ‘similarity’ of the two. Here, the number of nearest neighbours was set to 10.
\subsection{Decision Trees}
In the Decision Tree (DT) algorithm, the class of an input sample is predicted by moving from a root node to a leaf node. The path from root to leaf is determined by the decision criteria at each node, and the leaf node contains a specific output label. Decision Trees are developed in a top-down manner, and splitting is done at each node based on a certain decision criterion. This goes on until final leaf nodes are formed, representing the output classes \cite{Makhija2019}. The decision criteria for an input sample space is optimised during the training process. \\
\indent The Coarse tress allows a maximum of four splits for the decision-making process. This results in a small number of leaves to make class distinctions. In the Medium Tree, however, the maximum number of splits is 200, and more splitting points and leaf nodes are formed as a consequence.
\subsection{Discriminant Analysis}
The fundamental form of Discriminant Analysis is the Linear Discriminant Analysis (LDA). For a two-class problem with a single feature, the mean and covariance are calculated for each class. The LDA algorithm works on the assumption that all the involved classes have equal variances. A new axis is created in the feature space in such a way as to maximise the distance between the means of the two classes and minimise the variance of the dataset within each class. The data is then projected onto this axis, thereby separating the input space into two halves. Each of these halves represents a class. In situations where there is more than one feature, the covariance matrix is calculated and used in place of variance. Quadratic Discriminant Analysis is a variant of LDA, wherein an individual covariance matrix is determined for each of the classes. This increases the number of effective parameters and also accounts for the fact that the covariance matrices of each class may be different.
\subsection{Ensemble Trees}
Ensemble trees involves training several Decision Trees on the dataset, and then combining the output of all these trees in some way. It works by using the aggregate of decisions made by several trees. In the case of Bagged (Bootstrapping and Aggregating) trees, multiple sub-samples are pulled from the training data. A single DT is trained on each of these sub-samples, and then the aggregate of the outputs of each slightly different DT is used to make the final decision. This kind of ensemble learning has the advantage of combining several weak learners into a single stronger one.\\
\indent AdaBoosted Trees work by first training a DT on a random subset of the training data, giving each observation an equal weight. The performance of this tree is then evaluated. The second DT is trained by increasing the weights of the observations which the first DT had difficulty classifying, and lowering the weights of the observations which were easy to classify. After the second DT is trained, the combination of the two are evaluated, and the third DT is trained in a similar way. This continues and all of these learners are combined together to form a stronger learner. This kind of additive training ensures that subsequent trees classify observations that were not well classified by its predecessors.\\
\indent The RUSBoost (Random Undersampling Boost) method was designed specifically for training on imbalanced datasets\cite{Seiffert2010}. Random Undersampling is a technique that attempts to alleviate the class imbalance ratio by randomly removing instances of the majority class during training. The RUSBoost algorithm combines random undersampling with AdaBoost. This technique works the same way as the AdaBoost method described above, except that the majority class is randomly undersampled when selecting a sample set to train each successive DT. 

\subsection{Support Vector Machines}
 In Machine Learning, SVMs are one of the most useful tools used. Their ability to generalize well to unseen data allows them to achieve very high accuracies. SVMs have been known to perform well even in cases were the class imbalance ratio is as low as 1:10.  \\ 
 \indent A Support Vector Machine aims at finding a hyperplane that separates the two classes involved, while minimizing the distance (or margin) between both the classes. The hyperplane for a classification problem with $d$ features is a $(d-1)$ dimensional subset of the feature space that separates the two classes. The SVM method works by finding an optimal hyperplane that will maximize the margin between the hyperplane and the nearest sample. This hyperplane is a non-linear decision boundary in the original space, but can be construed to be linear in the higher dimensional input feature space \cite{Marwala2014}.

The training phase involves solving the Lagrangian \cite{Vapnik2000, Ng2012, Matasci2012} :
\begin{equation}
R(\vec{w},b) = \frac{(\vec{w}).(\vec{w})}{2} + C\sum_{i=1}^{l} \zeta_i , C >0
\end{equation}
Where $\vec{w}\) is a vector normal to the classifier line,
$C$ represents the tolerance to deviations, $l$ =  total no. of points and $\zeta_i$ represents the constraints placed on the system, namely
\begin{equation}
 \zeta_i = (1 - y_i(\vec{w}.\vec{x_i} + b))
\end{equation}
Here, $b$ is a constant, $y_i$ = (+1) for positive class samples and (-1) for negative class samples and $x_i$ is the $i$-th data point.
Using the condition mentioned in (2), (1) becomes

\begin{align}\label{1}
 R(\vec{w},b) = \frac{(\vec{w}).(\vec{w})}{2} + C \sum_{i=1}^{l} (1 - y_i(\vec{w}.\vec{x_i} + b)) 
\end{align}

Partially differentiating \eqref{1} w.r.t $\vec{w}\) and $b$ and equating the resulting terms to zero yields (4) and (5)
\begin{equation}
 \vec{w} = \sum_{i=1}^{l} (\alpha_i y_i x_i) 
\end{equation}
\begin{equation}
 \sum_{i=1}^{l} (\alpha_i y_i) = 0 
\end{equation}
Where the $\alpha_i$'s are the Lagrangian multipliers related to each training sample. Plugging the resulting conditions in \eqref{1} yields (6)
\begin{equation}\label{2}
R(\vec{w},b) = \sum_{i=1}^{1} (\alpha_i) - \frac{1}{2} \sum_{i=1}^{l}\sum_{j=1}^{l} (\alpha_i \alpha_j y_i y_j x_i x_j)
\end{equation}

Rather than applying SVMs directly on the input features $x_i$, a transform $\phi(x)$ is first performed. It can be seen that all the equations remain unchanged, apart from replacing $x_i.x_j$ with $\phi(x_i).\phi(x_j)$. Given a feature mapping function $\phi$, we define a \textbf{Kernel} as given by (7)

\begin{equation}
 K(x_i,x_j) = \phi(x_i).\phi(x_j) 
\end{equation}
The Linear Kernel is given by (8)
\begin{equation}
 K(x_i,x_j) = x_i^\top x_j 
\end{equation}
The Linear Kernel has the lowest training time of all other kernels, as only the value of $C$ needs to be optimised. With other kernel types like Polynomial, Gaussian, and Gaussian Radial Basis Function Kernels, an additional parameter, $\gamma$, needs to be optimised. The Polynomial Kernel is given by 

\begin{equation}
K(x_i,x_j) = (x_i^\top x_j + \gamma )^n 
\end{equation}

Where $n$ is the order of the Polynomial ($n$ is 2 for Quadratic and 3 for Cubic), and $\gamma \geq 0$ is a parameter that trades off the influence of higher order versus lower order terms in the polynomial.  

Another Kernel is the Gaussian Kernel given by (10). Here, $\gamma$ is a free parameter.
\begin{equation}
 K(x_i,x_j) = e^{ - \gamma \left( {x_i - x_j } \right)^2 } 
\end{equation}

\begin{figure}[htb!]
\begin{tikzpicture}[
box/.style={draw,rectangle,minimum size=1cm,text width=1cm,align=left}]
\matrix (conmat) [row sep=.1cm,column sep=.1cm] {
\node (tpos) [box,
    label=left:\( \mathbf{NQSO} \),
    label=above:\( \mathbf{NQSO} \),
    ] {1374};
&
\node (fneg) [box,
    label=above:\textbf{QSO},
    label=above right:\textbf{Total},
    label=right:\( \mathrm{1375} \)] {1};
\\
\node (fpos) [box,
    label=left:\( \mathbf{QSO} \),
    label=below left:\textbf{Total},
    label=below:1377] {3};
&
\node (tneg) [box,
    label=right:\( \mathrm{125} \),
    label=below:123] {122};
\\
};
\node [left=.05cm of conmat,text width=1.5cm,align=right] {\textbf{Actual \\ Class}};
\node [above=.05cm of conmat] {\textbf{Predicted class}};
\end{tikzpicture}
\caption{Confusion Matrix of LSVM during training}
    \label{fig:1}    
\end{figure}

\begin{figure}[htb!]
\begin{tikzpicture}[
box/.style={draw,rectangle,minimum size=1cm,text width=1cm,align=left}]
\matrix (conmat) [row sep=.1cm,column sep=.1cm] {
\node (tpos) [box,
    label=left:\( \mathbf{NQSO} \),
    label=above:\( \mathbf{NQSO} \),
    ] {3634};
&
\node (fneg) [box,
    label=above: \( \textbf{QSO} \),
    label=above right:\textbf{Total},
    label=right:\( \mathrm{3650} \)] {16};
\\
\node (fpos) [box,
    label=left:\( \mathbf{QSO} \),
    label=below left:\textbf{Total},
    label=below:3654] {20};
&
\node (tneg) [box,
    label=right:\( \mathrm{350} \),
    label=below:346] {330};
\\
};
\node [left=.05cm of conmat,text width=1.5cm,align=right] {\textbf{Actual \\ Class}};
\node [above=.05cm of conmat] {\textbf{Predicted class}};
\end{tikzpicture}
\caption{Confusion Matrix of LSVM on Test data}
    \label{fig:2}    
\end{figure}

\indent Given that missing even a single Quasar can be costly, even a 0.1\% variation in accuracy between methods is important. It can be seen in Table \ref{table:1} that the Linear SVM achieved the highest accuracy, and lowest misclassification cost. A misclassification cost of 4 indicates that a total of 4 objects were placed in the wrong category as can be seen in the confusion matrix. Five-fold cross-validation was used during training, wherein 5 disjoint test sets are used to test the model. The confusion matrix obtained during this training process is shown in Fig. \ref{fig:1}. \\
\indent On the remaining 40\% of the dataset left for testing, an accuracy of 99.1\% was achieved using the trained LSVM, with a False Negative Rate (FNR) of 5.71\% and a False Positive Rate (FPR) of 0.4\%. This can be observed in the confusion matrix in Fig. \ref{fig:2}. Even though an FNR of 5.71\% is relatively low, in an astronomical application, it is very high. Of the 350 Quasars in the test data, 20 were misclassified as Non-Quasar Objects. This represents a huge loss, given the rarity of these objects.

\section{Learning from Mistakes Technique}
 In order to decrease the FNR of the model, a Learning From Mistakes approach is used. As shown in \cite{Bhatnagar2010}, LFM techniques applied to SVMs perform well in cases of large imbalance in the minority and majority class ratio. This method involves cascading machine learning algorithms together to achieve an improved result. The misclassified data points of the first algorithm are used to train the next algorithm. In our case, the misclassified data points from the Linear SVM was used to train another set of ML algorithms. The outputs of both these models on a specific data sample are considered when making the final output class decision.\\
\indent To implement this method, the dataset was split in the ratio 2:1:1. The first 50\% of the data samples were used to train the LSVM, and an accuracy of 99.7\% was achieved while holding out 15\% for validation. This LSVM was then tested on the second 25\% of the data samples, and the misclassified samples were isolated by comparing with the target. There were a total of 23 misclassified objects (Quasars classified as Non-Quasars (10) and Non-Quasars classified as Quasars (13)), which were used to train the next set of algorithms.\\
\indent Two different validation schemes were used to train these algorithms. With a five-fold cross validation approach, the Ensemble Bagged Tree algorithm was found to be the best performing, and with 15\% holdout validation, another Linear SVM was found to be the best performing. Both models achieved an accuracy of 100\% during training. This is obviously skewed given the limited nature of the dataset used to train them, which comprised only of the samples misclassified by the first model.\\
 \indent To get the final output, both the sequentially trained algorithms are tested on the last 25\% of the data samples. A simple OR function is used to get the final predicted class. If either of the models predicted QSO, then the final class assigned was QSO as well. This effectively means that only samples classified as NQSO by both the models were placed in the NQSO class. The results are shown in the Confusion Matrices in Fig. \ref{fig:3} and Fig. \ref{fig:4}.

\begin{figure}[htb!]
\begin{tikzpicture}[
box/.style={draw,rectangle,minimum size=1cm,text width=1cm,align=left}]
\matrix (conmat) [row sep=.1cm,column sep=.1cm] {
\node (tpos) [box,
    label=left:\( \mathbf{NQSO} \),
    label=above:\( \mathbf{NQSO} \),
    ] {1213};
&
\node (fneg) [box,
    label=above:\textbf{QSO},
    label=above right:\textbf{Total},
    label=right:\( \mathrm{2286} \)] {1073};
\\
\node (fpos) [box,
    label=left:\( \mathbf{QSO} \),
    label=below left:\textbf{Total},
    label=below:1214] {1};
&
\node (tneg) [box,
    label=right:\( \mathrm{214} \),
    label=below:1286] {213};
\\
};
\node [left=.05cm of conmat,text width=1.5cm,align=right] {\textbf{Actual \\ Class}};
\node [above=.05cm of conmat] {\textbf{Predicted class}};
\end{tikzpicture}
\caption{Confusion Matrix of LSVM coupled with EBT}
    \label{fig:3}    
\end{figure}

\begin{figure}[htb!]
\begin{tikzpicture}[
box/.style={draw,rectangle,minimum size=1cm,text width=1cm,align=left}]
\matrix (conmat) [row sep=.1cm,column sep=.1cm] {
\node (tpos) [box,
    label=left:\( \mathbf{NQSO} \),
    label=above:\( \mathbf{NQSO} \),
    ] {7};
&
\node (fneg) [box,
    label=above:\textbf{QSO},
    label=above right:\textbf{Total},
    label=right:\( \mathrm{2286} \)] {2279};
\\
\node (fpos) [box,
    label=left:\( \mathbf{QSO} \),
    label=below left:\textbf{Total},
    label=below:7] {0};
&
\node (tneg) [box,
    label=right:\( \mathrm{214} \),
    label=below:2493] {214};
\\
};
\node [left=.05cm of conmat,text width=1.5cm,align=right] {\textbf{Actual \\ Class}};
\node [above=.05cm of conmat] {\textbf{Predicted class}};
\end{tikzpicture}
\caption{Confusion Matrix of LSVM coupled with LSVM}
    \label{fig:4}    
\end{figure}

\section{Performance Analysis Methodology}
 To analyse the performance of the trained model, several parameters are calculated using the testing data. These parameters are Precision, Sensitivity, Specificity, F–measure and g-mean \cite{Bhatnagar2010}. \\
\indent F-measure (also called F1 score) signifies the overall performance of the model. The F1 Score is the weighted harmonic mean of Precision and Sensitivity, and is a more all-encompassing measure of accuracy \cite{Indola2005}. G-mean (Geometric mean) signifies how well the classifier performs for both classes \cite{Bhatnagar2010}. The values of these performance parameters are shown in Table \ref{table:2}. 

 \begin{table}[htb!]
  \begin{center}
    \begin{tabular}{c|c|c|c|c|c} 
 Model & Precision  & Sensitivity   & Specificity  & F-measure   & g-mean \\ 
   & (\%) & (\%) & (\%) & (\%) & (\%) \\       
\hline LSVM & 95.38 & 94.29 & 99.56 & 94.83 & 96.89 \\ 
\hline \textbf{LSVM + EBT} & 16.63 & \textbf{99.53} & 53.06 & 28.5 & 72.67 \\ 
\hline LSVM + LSVM & 8.58 & 100 & 0.31 & 15.8 & 5.57 \\  
    \end{tabular}
  \end{center}
  \caption{Table of Coupled Model Performance}
  \label{table:2}
\end{table}

\section{Results}
\indent The False Negative Rate (FNR) with the addition of the Ensemble Tree was 0.4673\%, which represents a 10x reduction in FNR as compared to only one Linear Support Vector Machine (LSVM). This, however, comes at the cost of increasing the False Positive Rate (FPR) to 46.94\%. When two Linear SVMs were coupled together in this way, an FNR of 0\% was achieved, but with an extreme FPR of 99.69\%. This is impractical as this essentially means that almost all candidates were being classified as Quasars. 
The Sensitivity, which was 94.29\% for a single SVM trained on 6,000 data samples, increased to 99.53\% when a Linear SVM trained on 5000 samples was coupled with an Ensemble Bagged Tree algorithm. Thus, the sensitivity of the model to the minority class improved by more than 5\%, which in a dataset consisting of thousands of data points, represents a large number of correctly identified quasars. \\
\indent The increase in sensitivity when the LSVM is coupled with EBT comes at the cost of a high FPR. This is reflected clearly in the values of F-measure and g-mean, which both signify overall performance over both classes. This is why the F-measure and g-mean values are much lower for the coupled models than the single LSVM model.\\
\indent Even though the reduction in FNR comes at a cost of increasing the FPR, this is a reasonable trade-off to make while analysing astronomical data. It is much more preferable to follow up on false candidates than lose possible science targets for follow-up with specialised instruments. 
\section{Conclusion}
 With the growth of the data collected by land and space-based telescopes, machine learning will only become more relevant to deal with this influx of information. In this paper, a Learning From Mistakes approach was used to make a classifier to identify Quasars from the SDSS data. This approach has the advantage of drastically reducing the False Negative Rate (FNR) of the model, which is of paramount importance in Astronomy. It was observed that a Linear SVM together with an Ensemble Bagged Tree Algorithm had the best performance, in terms of FNR. This, however, comes at the expense of a high False Positive Rate (FPR). In most practical cases, it is necessary to analyse the false positives manually. If the cost of inspection is low enough, then this trade-off between FNR and FPR is viable.\\  
\indent This increase in FPR can be attributed to the fact that the training set for the second model is very limited. If the second Ensemble Trees model was trained with a larger dataset, there could be a reduction in the FPR as well, which is an option that can be explored in future research.

\bibliographystyle{ieeetran}
\bibliography{references2}

\end{document}